\documentclass[12pt]{article}

\usepackage{hyperref}
\usepackage{graphicx}
\usepackage{fullpage}
\usepackage{url}
\usepackage{tikz}
\usetikzlibrary{arrows,positioning,automata}
\usepackage{soul}
\usepackage{subfigure}
\usepackage{latexsym}	
\usepackage{xspace}
\usepackage{alltt}
\usepackage{bm}
\usepackage{amssymb}
\usepackage{amsmath}
\usepackage{xspace}
\usepackage{multirow}
\usepackage{times}
\usepackage{rotating}
\usepackage{listings}
\usepackage{caption}

\begin{document}

\begin{center}

{\large Towards Automated Safety Coverage and Testing for Autonomous Vehicles with Reinforcement Learning}

Hyun Jae Cho, and Madhur Behl \\
Department of Computer Science \\
University of Virginia \\
madhur.behl@virginia.edu
\end{center}

\section*{Abstract}
The kind of closed-loop verification likely to be required for autonomous vehicle (AV) safety testing is beyond the reach of traditional test methodologies and discrete verification. Validation puts the autonomous vehicle system to the test in scenarios or situations that the system would likely encounter in everyday driving after its release. These scenarios can either be controlled directly in a physical (closed-course proving ground) or virtual (simulation of predefined scenarios) environment, or they can arise spontaneously during operation in the real world (open-road testing or simulation of randomly generated scenarios).

In AV testing, simulation serves primarily two purposes: to assist the development of a robust autonomous vehicle and to test and validate the AV before release. A challenge arises from the sheer number of scenario variations that can be constructed from each of the above sources due to the high number of variables involved (most of which are continuous). Even with continuous variables discretized, the possible number of combinations becomes practically infeasible to test. To overcome this challenge we propose using reinforcement learning (RL) to generate failure examples and unexpected traffic situations for the AV software implementation. Although reinforcement learning algorithms have achieved notable results in games and some robotic manipulations, this technique has not been widely scaled up to the more challenging real world applications like autonomous driving.

The high level contribution of this research project is to build the framework for automatic scenario testing by using reinforcement learning. In particular, we use an AV simulator to try to cause a crash for an AV. A safe AV is expected to drive safely even in rare driving scenarios. We set up two simulations, one where the RL agent is given a full control over a set of variables, such as weather, position/velocity of an NPC vehicle, and time of day, to try to cause a crash for the AV. In the second version, the NPC vehicle follows a set of fixed waypoints, and the agent is to choose at which velocity it should be driving at each waypoint to cause a collision.

We discovered that in the first simulation, the reinforcement learning agent learned that setting a dense fog disabled the AV's perception module, which stopped it from moving in the middle of driving. In the second simulation, we noticed that the agent first let the AV follow its trajectory to its destination. When it arrived behind the NPC vehicle, the NPC vehicle traveled at a negative velocity, which the AV did not expect and resulted in a collision.

During our experiments, we use Baidu Apollo as our AV. We found out that in the first simulation, the RL agent sets a dense fog that invalidates Apollo's perception module from recognizing the traffic lights, making it unable to drive. In the second experiment, despite Apollo's robust AD stack, we were able to cause a direct collision between a non-autonomous NPC vehicle and Apollo. Therefore, we have revealed potential hazardous driving situations that current state-of-the-art autonomous vehicles are likely fail to drive safely. We believe that we have laid the framework for automated safety testing for autonomous vehicles. We hope that this way of safety testing will more widely be adopted in the future.

\pagenumbering{arabic}

\section{Background}
Today, countless tech corporations and startups are investing billions of dollars to develop autonomous vehicles (AV). In fact, several businesses will deploy their vehicles in the next several years~\cite{timeline}. Alphabet's autonomous vehicle business, Waymo, recently announced that their commercial self-driving service, Waymo One, will begin to deploy fully driverless cars~\cite{techcrunch}.

\begin{figure}
    \centering
    \includegraphics[width=\columnwidth]{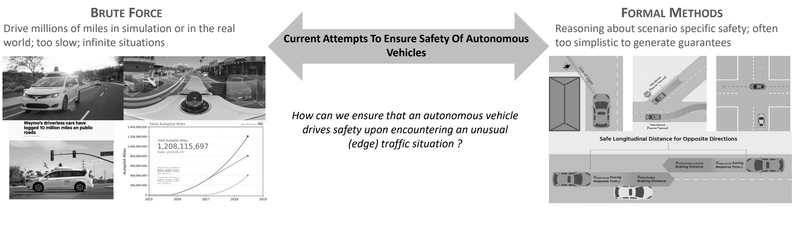}
    \caption{Traditional approaches to maneuvering autonomous vehicles rely on two broad methods: brute force and formal methods.}
    \label{fig:traditional}
\end{figure}

There are mainly two approaches in which traditional approaches for training autonomous vehicles have been taken, depicted in Figure \ref{fig:traditional}. First, autonomous vehicles are driven as many miles as possible in a brute force way to encapsulate as many different driving scenarios as possible. However, the innumerable combinations of the variables involved in driving, such as road geometry, vehicles, pedestrians, and weather, make it infeasible to encounter every possible scenario. For example, take Phantom AI's rear-end collision in 2018 \href{https://www.youtube.com/watch?v=zGoE6Hco4jE}{(link to video)}. Their L2 Advanced Driving Assistance System (ADAS) was unable to notice a dangerous cut-in by another vehicle and resulted in a crash. Another popular example is Uber's fetal crash that killed a pedestrian in 2018. 

\begin{figure}
    \centering
    \includegraphics[width=\columnwidth]{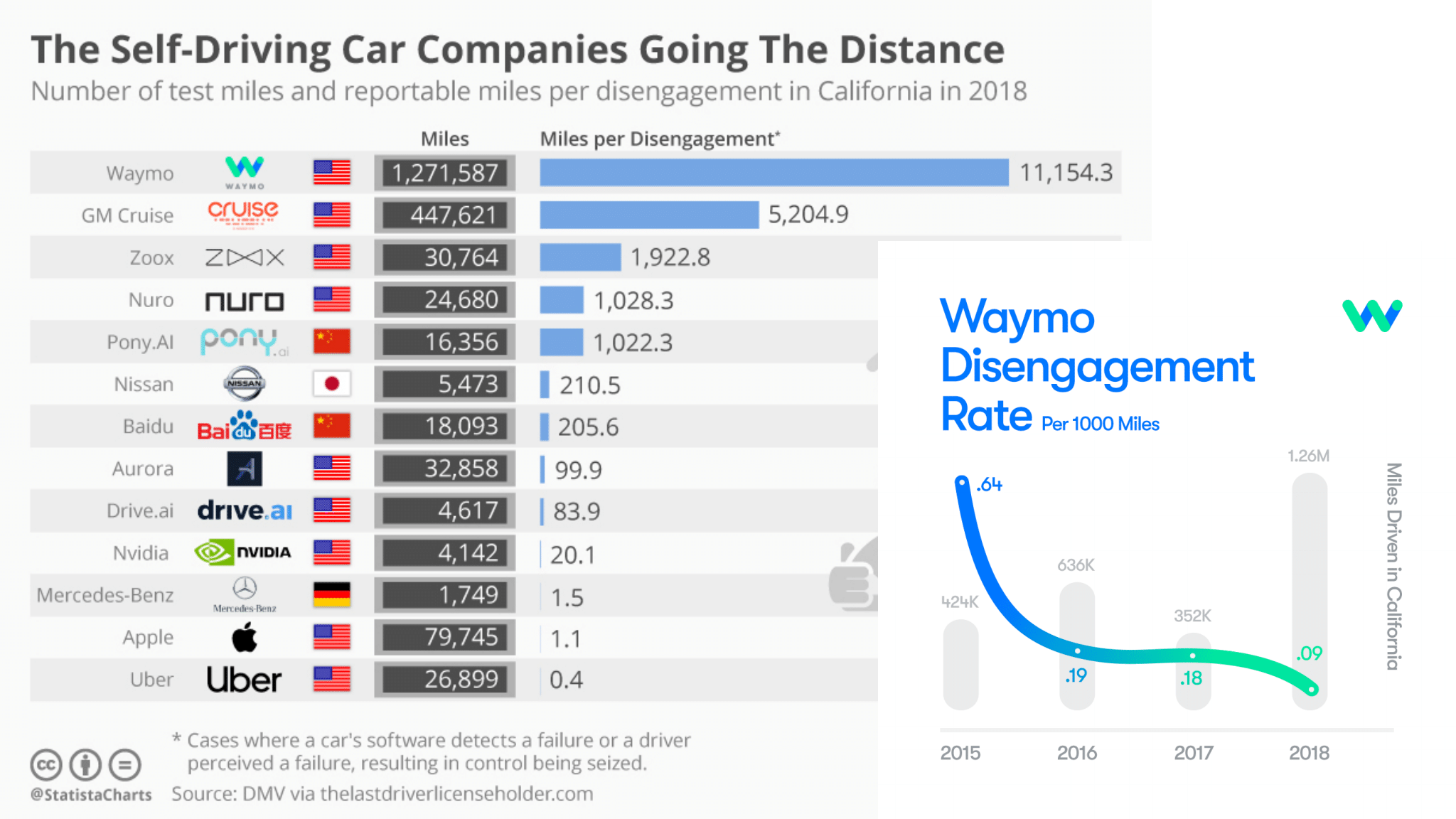}
    \caption{Autonomous driving companies such as Waymo drive many miles to encounter as many unique driving scenarios as possible. In fact, the number of disengagements has decreased substantially between 2015 and 2018. However, how much information does number of miles/disengagements tell about safety? Does this report--which is voluntarily assessed and reported by each company--necessarily imply that Waymo is safer than its competitors like GM Cruise or Baidu?}
    \label{fig:waymo}
\end{figure}

When companies advertise the safety of their autonomous vehicles by the number of miles they have driven, there is one important thing to consider: all companies drive their vehicles in different environments. An AV that drives the same high way a million times will not have been exposed to as many unique driving scenarios as another AV that drove 500 thousand miles in a crowded urban environment, such as New York City or Downtown Los Angeles. Therefore, the lack of unified safety testing method for AVs make it ambiguous to compare safety across multiple autonomous vehicles. For instance, a DMV report in 2018 shows that the number of human engagements in Waymo has decreased dramatically from 64\% in 2015 to 0.09\% in 2018, shown in Figure \ref{fig:waymo}. However, not all miles are equal since every company drives their AVs in different areas. These examples reveal that the brute force training approach to cover unique scenarios is just not sufficient. 

The other traditional approach for training autonomous vehicles is the use of formal methods, which reasons about specific driving scenarios and specifies scenario-specific driving behaviors. However, clearly formal methods is not capable of generating unique and more complicated scenarios. 

\section{Motivation}
The kind of closed-loop verification likely to be required for AV component testing is beyond the reach of traditional test methodologies and discrete verification.
Validation puts the autonomous vehicle system to the test in scenarios or situations that the system would likely encounter in everyday driving after its release. 
These scenarios can either be controlled directly in a physical (closed-course proving ground) or virtual (simulation of pre-defined scenarios) environment, or they can arise spontaneously during operation in the real world (open-road testing or simulation of randomly generated scenarios).

\begin{figure}
    \centering
    \includegraphics[width=\columnwidth]{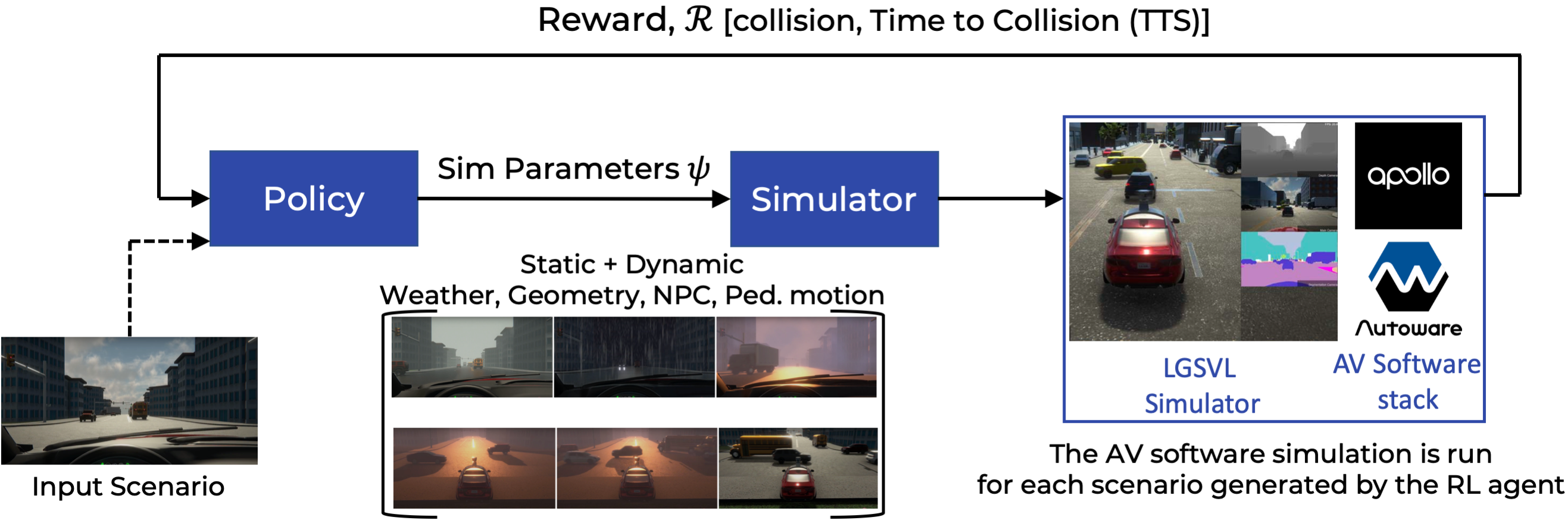}
    \caption{The reinforcement learning agent is tasked with the objective to maximize the reward associated with causing a collision for a given AV implementation. In this unique setting, the action space of the agent is the simulator/scene setup - i.e. the static and dynamic elements of the scene. After adequate iterations, the agent returns the scenario that is most likely to cause the AV to fail.}
    \label{fig:agent}
\end{figure}

In AV testing, simulation serves primarily two purposes: to assist the development of a robust autonomous vehicle and to test and validate the AV before release.
A challenge arises from the sheer number of scenario variations that can be constructed from each of the above sources due to the high number of variables involved (most of which are continuous). 
Even with continuous variables discretized, the possible number of combinations becomes practically infeasible to test. 
To overcome this challenge we propose using reinforcement learning (RL)~\cite{sutton2018reinforcement} to generate failure examples and unexpected traffic situations for the AV software implementation. 
Although RL algorithms have achieved notable results in games and some robotic
manipulations, this technique has not been widely scaled up to the more challenging real world applications like autonomous driving.

The frame of RL is an agent learning through interaction with its environment driven by an impact (reward) signal. The environment reinforces the agent to select new actions to improve learning process, hence the name of reinforcement learning~\cite{jaafra2018review}. 
RL algorithms have achieved notable results in many domains as games~\cite{mnih2015human,silver2016mastering} and advanced robotic manipulations~\cite{levine2016end, lillicrap2015continuous} beating human performance. However, standard RL strategies that randomly explore and learn lose efficiency and become computationally intractable when dealing with high-dimensional and complex environments~\cite{Wahlstrom2015FromPT}.
The origin of its difficulty lies in the important variability inherent to the driving task (e.g. uncertainty of human behavior, diversity of driving styles, complexity of scene perception, etc).

In our approach (Figure~\ref{fig:agent}), we rely on the LG Silicon Valley Lab (LGSVL) simulator~\cite{lgsvl}, a photo-realistic AV simulator that accepts scenario (or simulation) parameters through a Python API. 
For an input scenario, the API provides a way to vary the static (road geometry, position of traffic signs and lights, number of lanes, etc.) and dynamic parameters (weather, movement of other agents, pedestrians, etc.) of the simulator. The key insight is that the parameter space of the simulator, $\Psi$, is the action space for the reinforcement learning agent. 
For a given implementation of the AV software stack, such as Apollo from Baidu~\cite{fan2018baidu} or Autoware~\cite{kato2015open}, the agent's objective is to choose a set of simulation parameters (static and dynamic) such that it can cause the AV to fail. For each instance of the simulated scenario, the agent receives a reward if it can cause the AV to crash. Additionally, it receives a continuous reward to minimize a safety metric, such as time-to-collision.
The learned policy is a set of simulation parameters that could cause the AV to fail on the chosen safety metric. 

\section{Related Work}
To overcome the shortcomings of the two traditional methodsof AV training, several recent studies were conducted to train AVs under unexpected, rare scenarios using reinforcement learning agents and simulation. First, Wang\textit{ et al}.~\cite{Wang_2018} applied reinforcement learning to train autonomous vehicles for lane changing tasks in unexpected scenarios. When changing the lane, the lag vehicle in the target lane can take multiple different actions, such as yielding by decelerating, accelerating to prevent cut-in, or avoid by changing lane. To explore the variability, they implemented a model-free reinforcement learning approach to find the optimal policy when changing lanes. In their simulation, they includeed some randomness in the parameters, such as the departure interval and individual speed limits for vehicles. However, their model was trained on a fixed traffic geometry: a highway segment with fixed number of lanes, length and width. Therefore, their approach is not scalable to other road conditions.

In addition, Sarkar \textit{et al}~\cite{sarkar2019behavior} used the bounded rationality theory to develop a driving behavior model which generates rate driving scenarios. Their model used existing driving data to efficiently generate rare driving scenarios that are strongly correlated to naturalistic driving data. Furthermore, O'Kelley \textit{et al}.~\cite{okelly2018scalable} developed a simulation framework that can test an entire autonomous driving system. Furthermore, they used adaptive importance-sampling to evaluate rare-event probability and the probability of an accident in a simulated driving scenario.

To our knowledge, there has not been a study on using reinforcement learning and an autonomous vehicle simulator to explicitly target collisions on existing AVs such as Baidu's Apollo, which is the goal of this research project.

\section{Problem Statement}
Let $S$ represent a simulation scenario. In $S$, the RL agent can control a set of parameters $\Psi$, some of which are dynamic $\Psi_{d}$ and others static $\Psi_{s}$. 

\begin{figure}
    \centering
    \includegraphics[width=0.5\columnwidth]{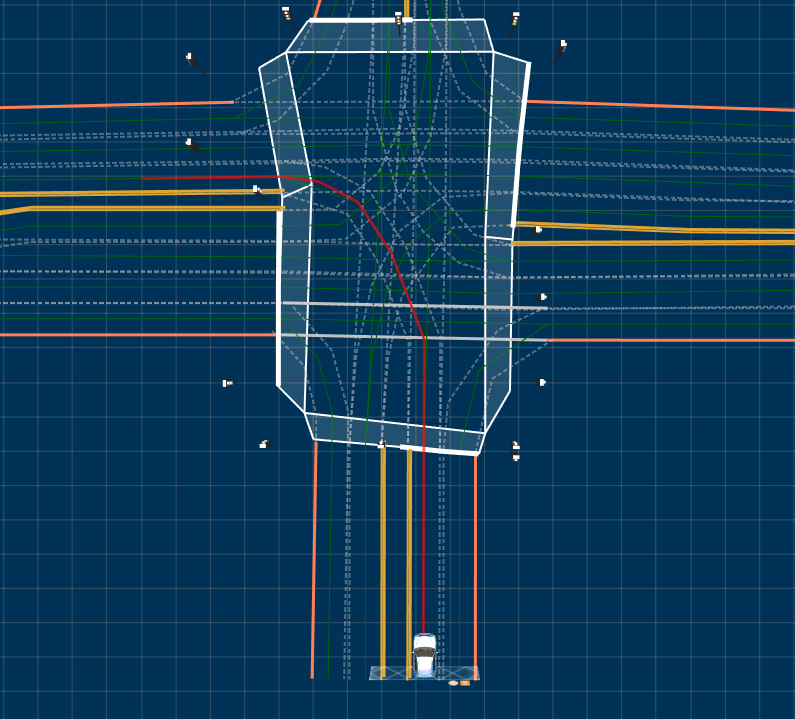}
    \caption{The AV is spawned at the bottom. It follows a trajectory (red) until it reaches its destination.}
    \label{fig:routing}
\end{figure}

We use RL to generate two simulations: $S_{1}$ and $S_{2}$. In both $S_{1}$ and $S_{2}$, we spawn the AV at the same location and select the same destination point (Figure \ref{fig:routing}). The NPC starts from the opposite side of the crossroad and drives towards the AV. Therefore, the targeted location of collision is where the trajectories of the two vehicles coincide.

The reward function for the RL agent relies on time-to-collision (TTC) between the NPC vehicle and the Apollo AV. $$TTC(NPC, AV) = \frac{distance(NPC, AV)}{|speed(NPC) - speed(AV)|}$$ The denominator is the absolute value of the relative speed of the AV to NPC's. As the distance reduces and the two vehicles drive faster toward each other, TTC will decrease. Therefore, the reward function $R(t)$ for time step $t$ is defined as 

\[R(t) = \begin{cases} 
      100 &\text{if collision occurs} \\
      \frac{1}{TTC(NPC, AV)} &\text{otherwise}
   \end{cases}
\]

This reward function encourages a collision because the agent can gain higher reward by reducing TTC and by causing a collision. A collision reward of 100 was chosen ineptly.

\section{Reinforcement Learning Background}
\begin{figure}
    \centering
    \includegraphics[width=0.6\columnwidth]{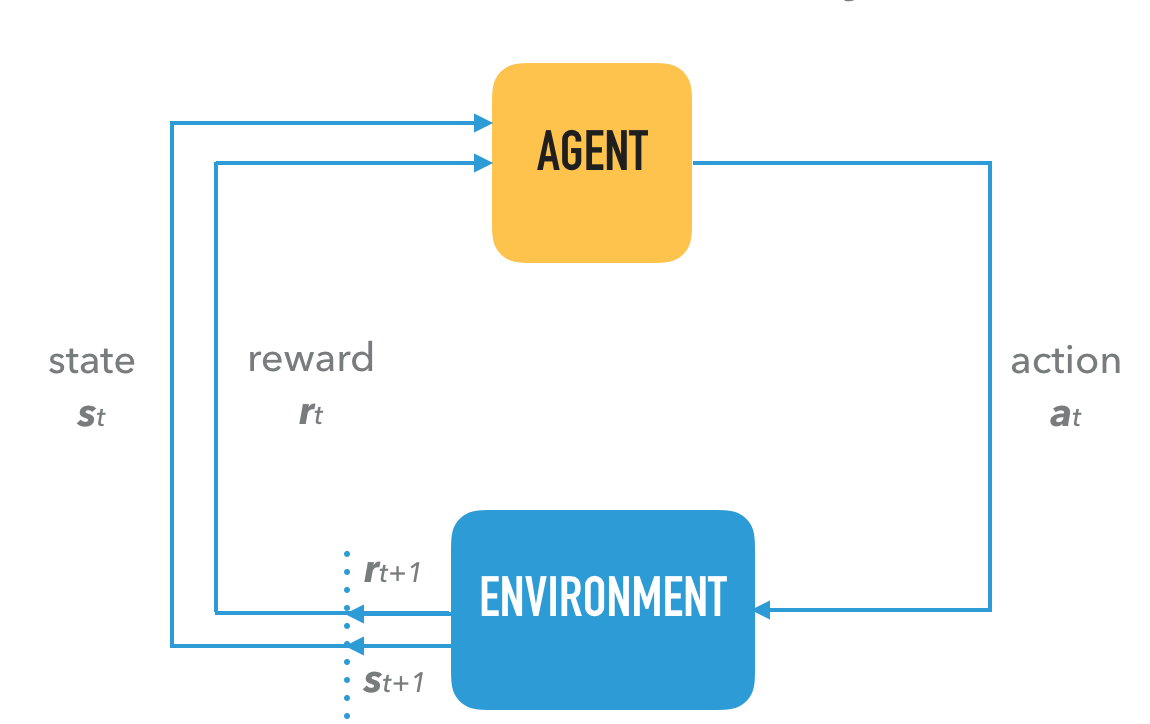}
    \caption{Diagram of reinforcement learning~\cite{rlbook}. At time step $t$, the agent takes action $a_{t}$. The environment responds by outputting the new state $r_{t+1}$ and reward $S_{t+1}$. This cycle continues until the end of training the model.}
    \label{fig:rlarchitecture}
\end{figure}
Reinforcement learning (RL) is an unsupervised learning algorithm. The framework in RL involves five main parameters: environment, agent, state, action, and reward. The environment is the world in which the agent moves. Due to the unsupervised nature of RL, the agent does not start out knowing the notion of good or bad actions. At each time step, it performs an action in the environment that leads to the next state and the corresponding reward (Figure \ref{fig:rlarchitecture}). As training progresses, the agent learns which are good and bad actions by how much reward it can obtain from each action. Therefore, the reward function plays a vital role for the agent's ability to learn a policy--a set of actions--that leads to good performance.

\begin{figure}
    \centering
    \includegraphics[width=0.7\columnwidth]{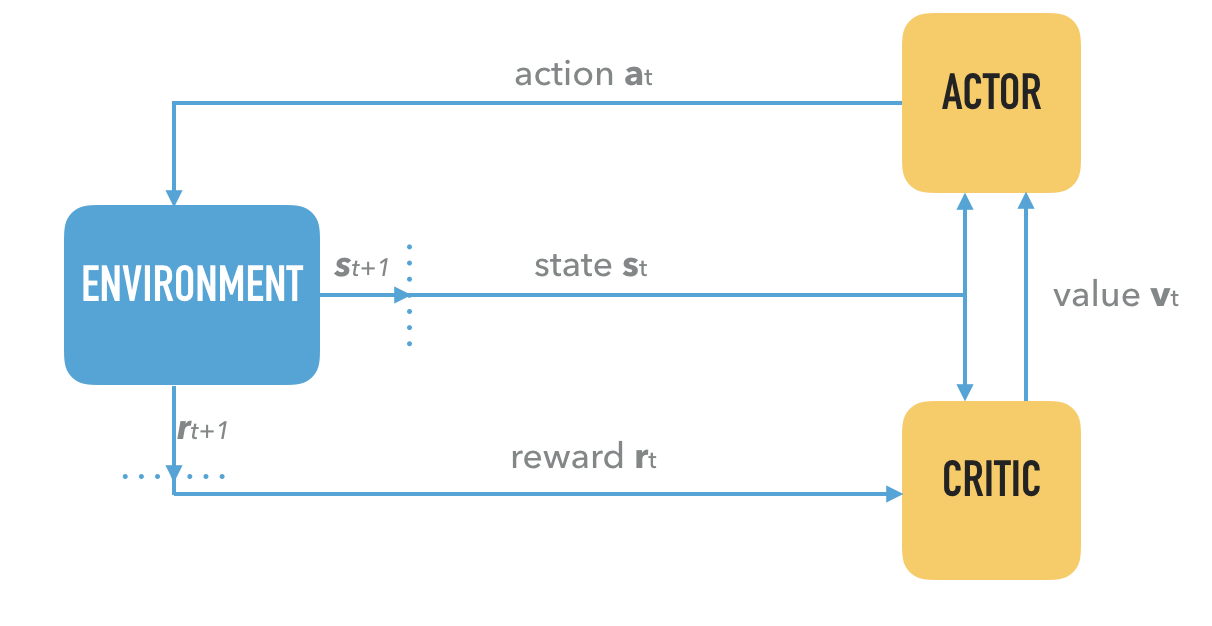}
    \caption{Architecture of actor critic model~\cite{dlbook}. At time step $t$, the agent takes action $a_{t}$. The environment responds by outputting the new state $S_{t+1}$ and reward $r_{t+1}$. This cycle continues until the end of training the model.}
    \label{fig:ac}
\end{figure}

Actor-critic (Figure \ref{fig:ac}) is a reinforcement learning algorithm that incorporates two models. The actor model outputs the desired action at each state. The critic model takes in that action to estimate its value--the goodness of being in that state. Usually, each model implements a neural network. At each time step, the weights of the actor model are updated to perform actions to maximize the value, which the critic model predicts. Therefore, as training progresses, the actor becomes better at making good actions and the critic becomes better at estimating the value in each state. Please refer to pseudo code above for the step by step process.

\begin{lstlisting}[title=Actor Critic Algorithm., float, frame=tb]
Initialize $V(s)$, the expected reward at state $s$, arbitrarily
and $\pi$ to the policy to be evaluated.
For each episode,
    Initialize s.
    $a$ $\leftarrow$ action given by actor.
    Take action $a$. Observe reward $r$, and next state $s^\prime$.
    $V(s) \leftarrow V(s) + \alpha[r + \gammaV(s^\prime)-V(s)]$
    $s \leftarrow s^\prime$.
until $s$ is terminal.
\end{lstlisting}

\section{Reinforcement Learning for Safety Coverage Testing}
In this section, we describe in detail our process of applying RL to safety coverage testing. Specifically, we describe the LGSVL simulator, Baidu Apollo's AD stack (perception, planning, and control modules), and the integration of Apollo on the simulator. Finally, we discuss our formulation of RL to Apollo, which is implemented using the simulator's Python API.

\subsection{The LGSVL Simulator}
The LGSVL simulator is an open-source, Unity-based autonomous vehicle simulator. It allows AV developers to test their AD algorithms. Currently, it supports Autoware and Apollo 3.5 and 5.0 platforms. In addition to the provided maps and vehicles right out of installation, the simulator allows creating, editing, and annotating HD maps and vehicles for users.

\begin{figure}
    \centering
    \includegraphics[width=0.75\columnwidth]{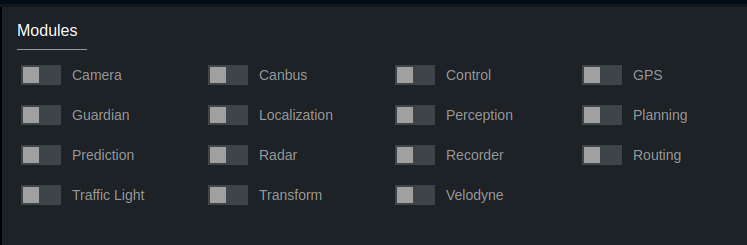}
    \caption{In our case, we follow the guideline by LGSVL and enable localization, perception, planning, prediction, control, routing, traffic light, and transform.}
    \label{modules}
\end{figure}

\begin{figure}
\centering
\begin{minipage}[t]{.5\textwidth}
  \centering
  \includegraphics[width=6cm, height=5cm]{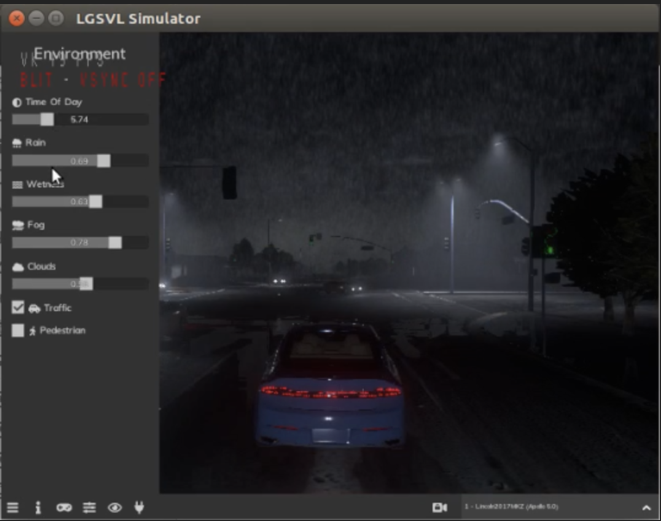}
    \captionsetup[figure]{width=.85\textwidth}
  \captionof{figure}{The Python API provides an array of functionalities that can make desired scenarios. Change and visualize  time of day, weather traffic.}
  \label{pyAPI1}
\end{minipage}%
\begin{minipage}[t]{.5\textwidth}
  \centering
  \includegraphics[width=6cm, height=5cm]{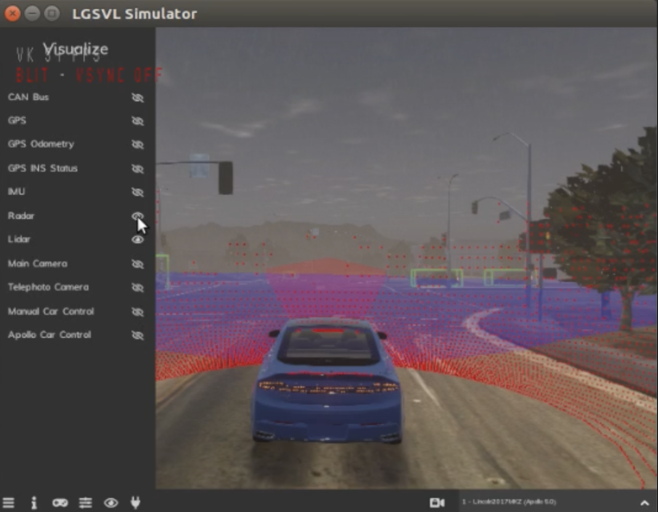}
    \captionsetup[figure]{width=.85\textwidth}
  \captionof{figure}{Visualize sensors and bounding boxes.}
  \label{pyAPI2}
\end{minipage}
\end{figure}

\begin{figure}
\centering
\begin{minipage}[t]{.5\textwidth}
  \centering
  \includegraphics[width=6cm, height=5cm]{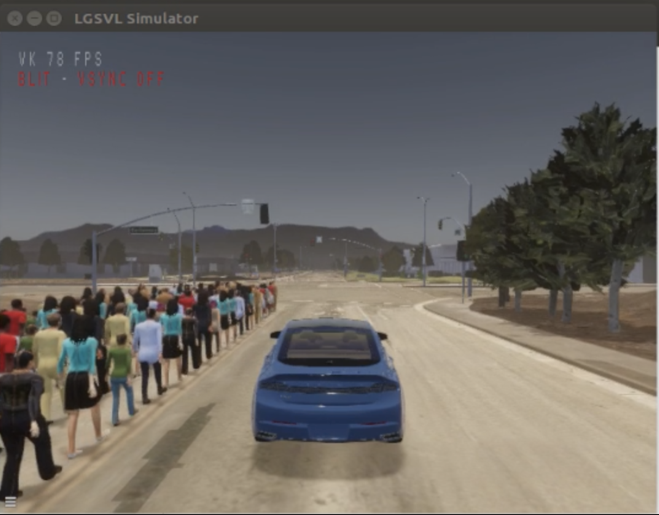}
  \captionsetup[figure]{width=.85\textwidth}
  \captionof{figure}{Spawn and give waypoints to NPC pedestrians.}
  \label{pyAPI3}
\end{minipage}%
\begin{minipage}[t]{.5\textwidth}
  \centering
  \includegraphics[width=6cm, height=5cm]{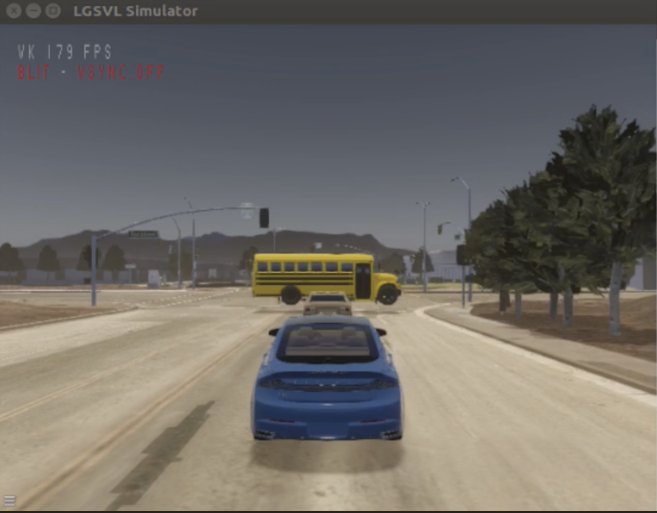}
  \captionsetup[figure]{width=.85\textwidth}
  \captionof{figure}{Spawn NPC vehicles at desired locations.}
  \label{pyAPI4}
\end{minipage}
\end{figure}

The simulator provides a web user interface (Dreamview) that communicates with Apollo on simulator via rosbridge. On Dreamview, users can perform a variety of tasks, including spawning NPC vehicles and pedestrians, visualizing sensors, and changing weather parameters, including rain, fog, and wetness. The locations and types of NPC vehicles and pedestrian spawns are chosen randomly. Using the Python API, NPCs and other parameters can be deterministic (Figures \ref{pyAPI1}, \ref{pyAPI2}, \ref{pyAPI3}, \ref{pyAPI4}). Users can turn on the modules in Figure \ref{modules} to enable each functionality for the vehicle.

We used LGSVL's BorregasAve map, which renders an actual location in Sunnyvale, CA. We use the September 2019 release of the simulator.

\subsection{Baidu Apollo's AD Stack}

\begin{figure}
\centering
  \centering
  \includegraphics[width=\columnwidth]{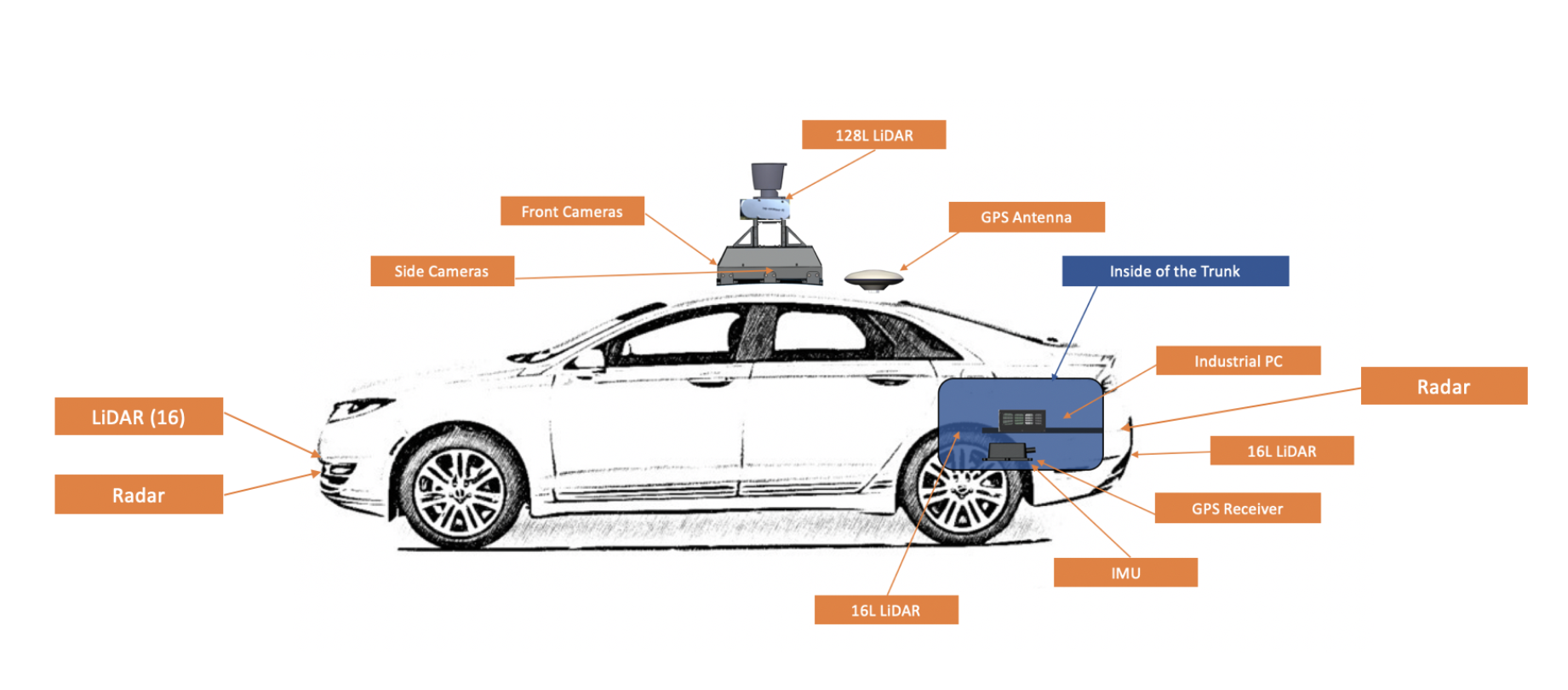}
  \caption{Hardware architecture of Baidu's Apollo. Adequate hardware, including LiDAR, radar, GPS, and IMU, are equipped to enable the AD stack.}
  \label{apollohardware}
\end{figure}

\begin{figure}  \centering
  \includegraphics[width=0.8\columnwidth]{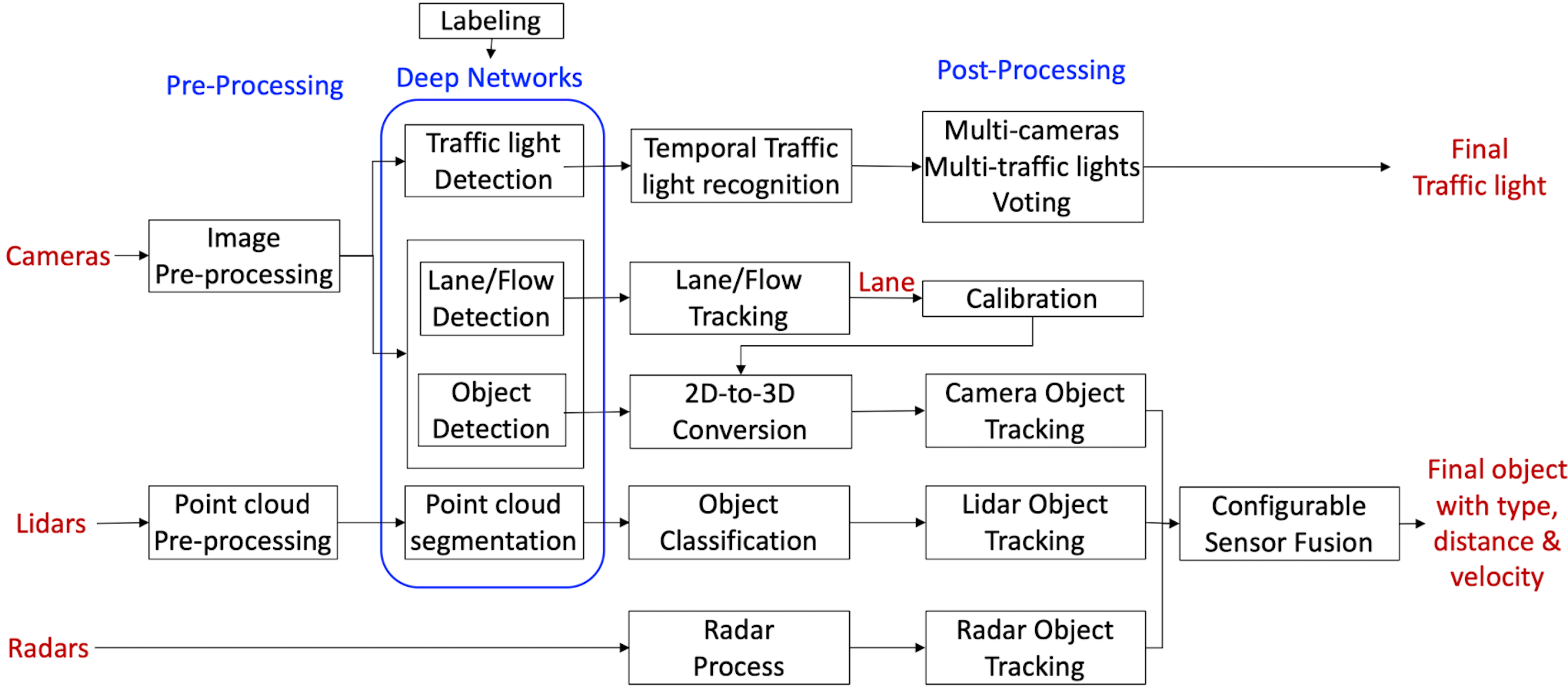}
  \captionsetup[figure]{width=.85\textwidth}
  \captionof{figure}{Apollo's perception pipeline. Apollo uses deep neural networks on combined data from multiple sensors, such as camera, LiDAR, and radar, for traffic light and object detection.}
  \label{apolloperception}
\end{figure}

Perception, planning, and control are often referred to as the autonomous driving stack (AD stack). Perception is the module that enables AVs to make sense of the surroundings. Apollo uses camera images, which are processed and fed into YOLO v2~\cite{yolo2} object detection algorithm, to locate and classify pedestrians, traffic signs, lights, other vehicles, and other objects. Then, it accurately locates itself around the surrounding objects by using a combination of data from IMU, GPS, radar, and LiDAR, which is then compared to a previously annotated high definition map (HD map) for increased precision. Refer to Figure \ref{apolloperception} for a detailed pipeline of how Apollo combines data from multiple sources for detecting traffic lights and objects.

Apollo's planning module generates a trajectory defined by a series of waypoints. It calculates the time and velocity it should be driving at when passing through each waypoint. A curve is then fit on the waypoints to produce a smooth representation of the trajectory. Planning also observes the movements of other vehicles and adjusts its waypoints and velocity accordingly in order to avoid collisions.

\begin{figure}
    \centering
    \includegraphics[width=0.5\columnwidth]{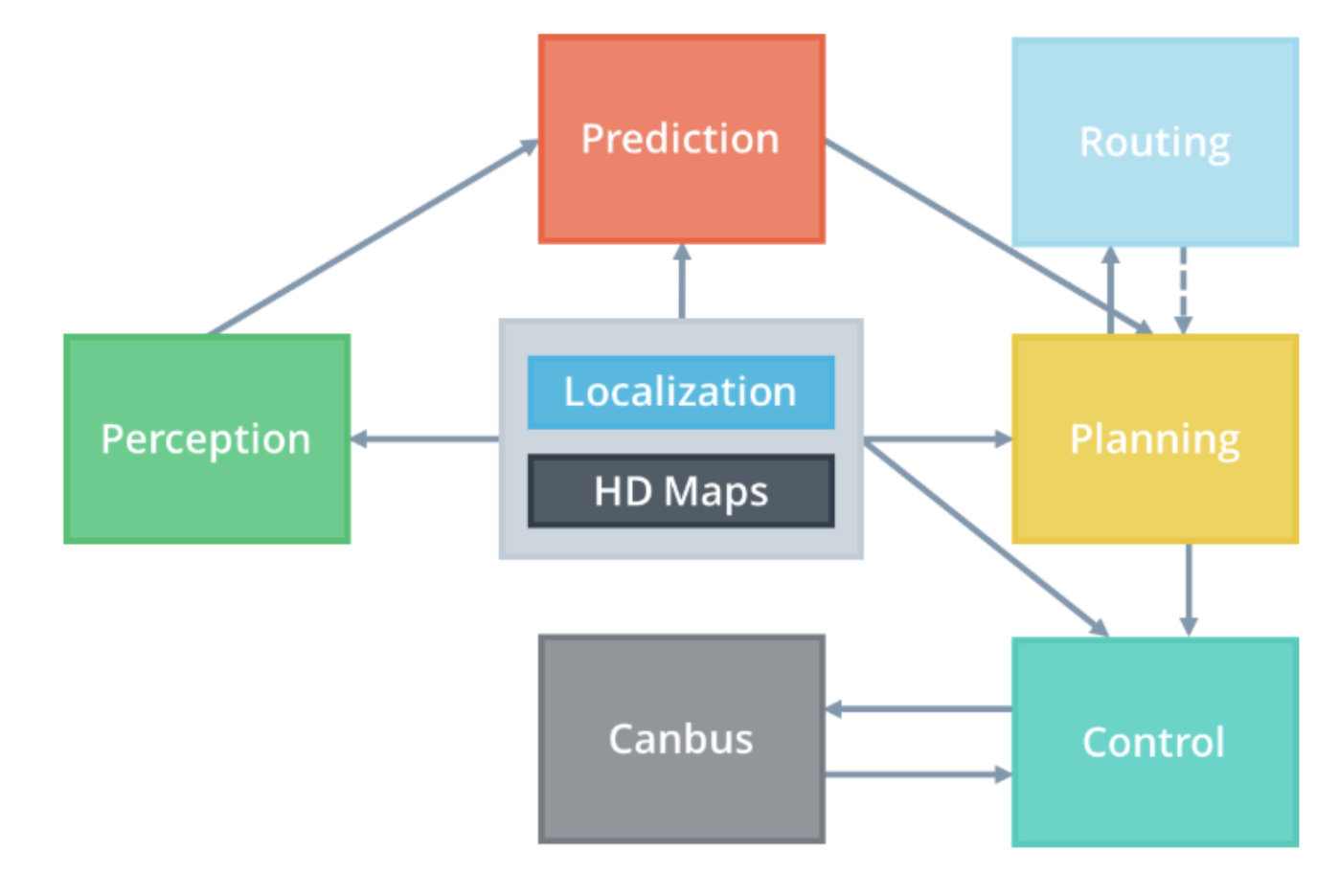}
    \caption{Diagram showing how the modules from HD maps/localization to CAN bus work together in Apollo. CAN bus is a communication protocol that allows components within a vehicle seamlessly. Image source: Udacity's Self-Driving Fundamentals: Featuring Apollo course.}
    \label{fig:apollomodules}
\end{figure}

The control module is what actually moves the vehicle by controlling acceleration, braking, and steering angle of Apollo. The waypoints and corresponding velocities generated from the planning module are given as the input to this module (Figure \ref{fig:apollomodules}). Apollo uses both proportional integral derivative (PID) controller and linear quadratic regulator (LQR) to control the degree of change in acceleration, braking, and steering angle. PID is simpler to calculate, but can overshoot the reference trajectory. LQR complements PID by using the state of the vehicle to minimize lateral error along the trajectory.

Apollo contains appropriate hardware to support the above modules, shown in Figure \ref{apollohardware}.

\subsection{Integration of Apollo on the LGSVL Simulator}
Apollo receives waypoints from Dreamview via rosbridge and transfers control commands (accelerate, brake, steering angle) to the simulator which moves the vehicle. The provided Python API allows sending and receiving websockets to and from Dreamview to change $\Psi_{d}$, which consists of status of traffic lights, weather, NPC vehicles, and pedestrians.

\subsection{RL Formulation to Apollo}
We can formulate RL to conduct safety coverage testing of AVs in a simulated scenario $S$. In every $S$, the Apollo vehicle (EGO) is spawned at a fixed location and is given a destination with its AD stack on. Therefore, the map (Borregas Ave), $\Psi_{s}$ (road geometry, position of traffic signs and lights, number of lanes) and EGO together form the environment of each scenario. $\Psi_{d}$ form the agent. An agent's action is a change in $\Psi_{d}$ at each time step $t$. Finally, the reward function $R(t)$ is set to the inverse function of time-to-collision (TTC) between the NPC vehicle and EGO in order to maximize the likelihood of crashing.

The actor of the RL model directs actions for the agent by calculating how much to change each of $\Psi_{d}$. The critic then estimates the value of each action by the agent.

\section{Simulation Results}
We have designed our experiment in two versions. First, we allowed the RL agent to change the following $\Psi_{d}$ in each time step: the position and velocity of one NPC vehicle, time of day, as well as the amount of rain, fog, and wetness (continuous values in [0.0, 1.0]). Therefore, the agent's action space included changing any combination of these variables. In the second version, we put restrictions on the agent such that it was only able to change the velocity of the NPC vehicle. We uniformly sampled a fixed number of waypoints so that the NPC reached them at each time step. In this version, the agent's job was to figure out the velocity to reach each sampled waypoint to increase the chance to crash with the EGO vehicle. However, the agent was allowed to travel at negative velocity, in which case it would move backwards.

\begin{figure}
\centering
\begin{minipage}[t]{.5\textwidth}
  \centering
  \includegraphics[width=7cm, height=5cm]{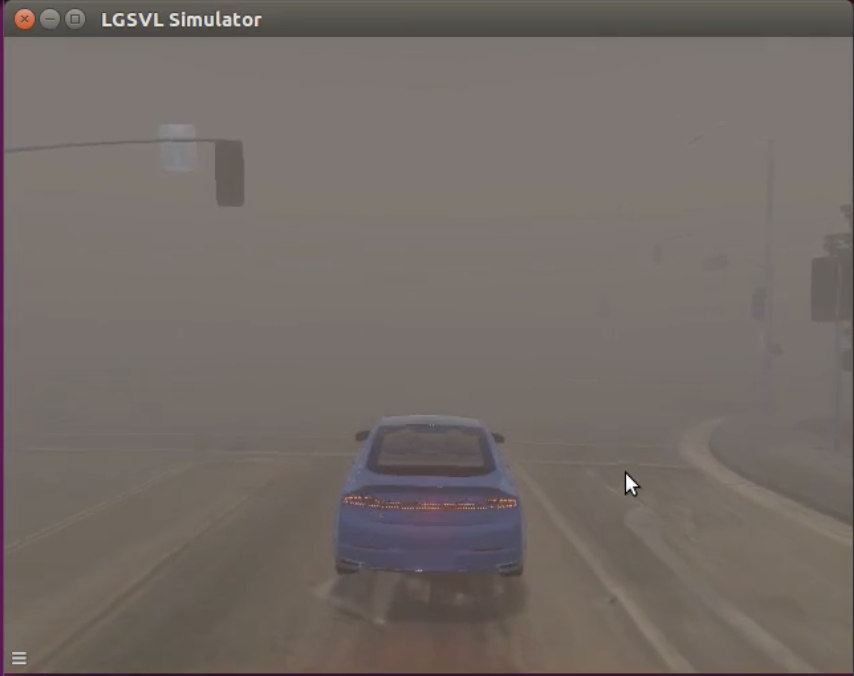}
  \captionsetup[figure]{width=.85\textwidth}
  \captionof{figure}{The RL agent has set a dense fog (fog rate=100\%).}
  \label{fog0}
\end{minipage}%
\begin{minipage}[t]{.5\textwidth}
  \centering
  \includegraphics[width=7cm, height=5cm]{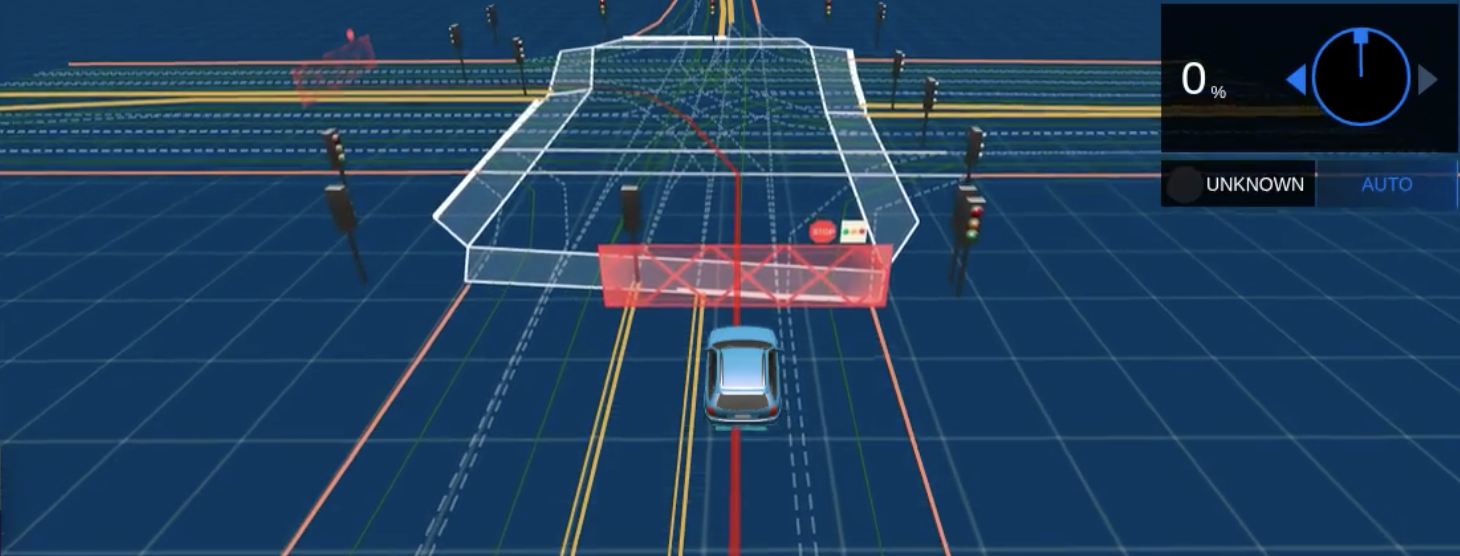}
  \captionsetup[figure]{width=.85\textwidth}
  \captionof{figure}{Dreamview of the simulator in Figure \ref{fog0}. The EGO vehicle is unable to read the traffic light, indicated by the "UNKOWN" sign.}
  \label{fog1}
\end{minipage}
\end{figure}

\subsection{Experiment 1}
We discovered that the RL agent will set a dense fog (Figure \ref{fog0}) to invalidate Apollo’s perception stack from properly perceiving the traffic light (Figure \ref{fog1}), yielding Apollo to stop driving. Although we believe that it is a logically correct behavior for Apollo to not move, it is also possible that suddenly stopping in the middle of driving will bring detrimental results of colliding with other vehicles, which might be moving. This result unveils the potential inaccuracy of the conventional safety metrics (miles-per-intervention and the total number of miles a self-driving car has driven) because we were able to generate a scenario where Apollo’s behavior was not optimum.

\begin{figure}
\centering
\begin{minipage}[t]{.5\textwidth}
  \centering
  \includegraphics[width=6.5cm, height=5cm]{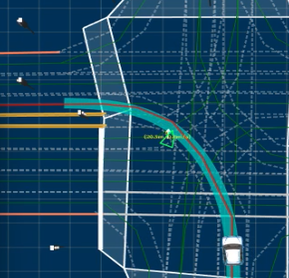}
  \captionsetup[figure]{width=.85\textwidth}
  \captionof{figure}{Dreamview of EGO following its trajectory toward its destination. Note that the green polygon in the path represents the existence of the NPC vehicle.}
  \label{npcin}
\end{minipage}%
\begin{minipage}[t]{.5\textwidth}
  \centering
  \includegraphics[width=6.5cm, height=5cm]{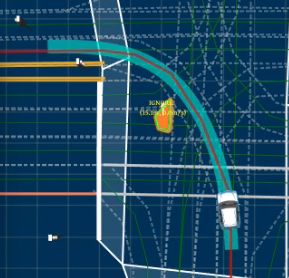}
  \captionsetup[figure]{width=.85\textwidth}
  \captionof{figure}{The NPC vehicle has moved out of the EGO vehicle's path and is marked as "Ignore".}
  \label{npcout}
\end{minipage}
\end{figure}

\subsection{Experiment 2}
\begin{figure}
\centering
\begin{minipage}[t]{.5\textwidth}
  \centering
  \includegraphics[width=6.5cm, height=5cm]{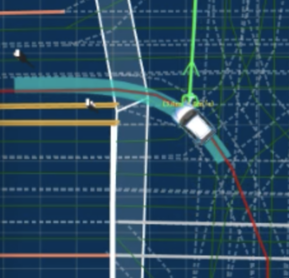}
  \captionsetup[figure]{width=.85\textwidth}
  \captionof{figure}{The NPC vehicle and Apollo collide. Apollo realizes that the NPC vehicle is driving backwards, indicated by the green arrow, but it was too late.}
  \label{npccrash}
\end{minipage}%
\begin{minipage}[t]{.5\textwidth}
  \centering
  \includegraphics[width=6.5cm, height=5cm]{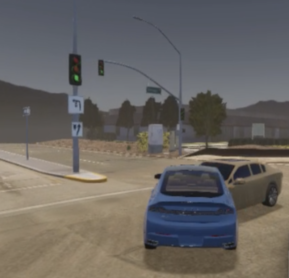}
  \captionsetup[figure]{width=.85\textwidth}
  \captionof{figure}{\ref{npccrash} in the LGSVL simulator view.}
  \label{npcrash2}
\end{minipage}
\end{figure}

When a moving object, such as an NPC vehicle or pedestrian, is on the trajectory, Apollo will stop and wait until the trajectory is clear of any obstacle. In Figure \ref{npcin}, Apollo has perceived the existence of an NPC vehicle (green polygon) on its path. However, as it drove closer, the NPC moved out of the trajectory. This led Apollo to believe that the NPC was traveling to the south. Hence, Apollo kept following its trajectory without stopping. However, the NPC vehicle did not keep driving south; it waited until Apollo was right behind (Figure \ref{npcout}) and drove at negative velocity, resulting in a crash (Figures \ref{npccrash}, \ref{npcrash2}). 

\section{Limitations}
The main computational bottleneck in training the RL model came from the LGSVL simulator. Since Apollo required the camera images to enable its perception stack, each scenario had to be rendered on the simulator, delaying the time that could have been spent on exploring more diverse scenarios. Even having parallel computing would not help because of the rendering problem. The lack of exclusive command line method of training was the main limitation in this project. 


\section{Conclusion and Future Work}
In this project, we used actor critic reinforcement learning to find edge cases for Baidu Apollo's AD stack using the LGSVL simulator. We successfully generated two edge cases. In the first case, the RL agent controlled a set of $\Psi_{d}$, such as weather, one NPC vehicle's positions and velocities, and time of day to cause a collision between the NPC vehicle and Apollo. Although it did not specifically crash the two vehicles, the agent impaired the perception module of Apollo from functioning correctly by creating a dense fog. Our second edge case, the NPC vehicle was given a set of waypoints, and agent's job was to figure out at which velocities it should travel each waypoint to crash with Apollo. In this scenario, the NPC reached until certain waypoint, where it stopped and drove backwards when Apollo was behind. The fact that we were able to generate a collision is significant; it reveals that the AVs today are potentially vulnerable to unexpected driving scenarios. We hope that further studies on automated safety testing for autonomous vehicles can be performed to reveal any other potentially dangerous situations.

Future work can include refining the reward function, adding more NPC vehicles and pedestrians, and extending the action space of the RL agent.

\newpage
\bibliographystyle{unsrt}
\bibliography{references}

\end{document}